\documentclass[titlepage,oneside,letterpage,12pt]{article}

\usepackage[tiny,rm]{titlesec}
\newpagestyle{trbstyle}{
\sethead{Wu, Balkumar and Luo et al.}{}{\thepage}
}
\titleformat{\section}{\bfseries}{}{0pt}{\uppercase}
\titlespacing*{\section}{0pt}{12pt}{*0}
\titleformat{\subsection}{\bfseries}{}{0pt}{}
\titlespacing*{\subsection}{0pt}{12pt}{*0}
\titleformat{\subsubsection}{\itshape}{}{0pt}{}
\titlespacing*{\subsubsection}{0pt}{12pt}{*0}

\usepackage{enumitem}
\setlist[1]{labelindent=0.5in,leftmargin=*}
\setlist[2]{labelindent=0in,leftmargin=*}

\usepackage[font={bf},tableposition=top,figureposition=bottom]{caption}
\usepackage{amsmath}
\makeatletter
\renewcommand{\fnum@figure}{\textbf{FIGURE~\thefigure} }
\renewcommand{\fnum@table}{\textbf{TABLE~\thetable} }
\makeatother

\usepackage{mathptmx}

\usepackage[T1]{fontenc}
\usepackage{textcomp}

\usepackage[sort,numbers]{natbib}

\setcitestyle{round}

\usepackage[pagewise]{lineno}

\newread\somefile
\usepackage{xparse}

\newcounter{totalwordcounter}
\newcounter{wordcounter}
\makeatletter

\NewDocumentCommand{\wordcount}{s}{%
  \immediate\write18{texcount -sum -1 \jobname.tex > count.txt}%
  \immediate\openin\somefile=count.txt%
  \read\somefile to \@@localdummy%
  \immediate\closein\somefile%
  \setcounter{wordcounter}{\@@localdummy}%
  \IfBooleanF{#1}{%
  \@@localdummy
  }%
}
\makeatother

\usepackage{totcount}
	\regtotcounter{table} 	
	\regtotcounter{figure} 	

\newcommand{\wordfigure}{250} 
\newcommand{\wordtable}{250} 

\newcommand{\totalwordcount}{%
  \wordcount*
  \setcounter{totalwordcounter}{\value{wordcounter}}%
  \addtocounter{totalwordcounter}{\numexpr\wordfigure*\totvalue{figure}}%
  \addtocounter{totalwordcounter}{\numexpr\wordtable*\totvalue{table}} %
  \number\value{totalwordcounter}
  \renewcommand{\totalwordcount}{\number\value{totalwordcounter}}
}

\usepackage{float}
\usepackage{graphicx}

\usepackage{nomencl}
\makenomenclature

\usepackage{algorithm}
\usepackage{algpseudocode}

\usepackage{verbatim}
\usepackage{subfig}
\usepackage{url}

\begin{document}
\captionsetup[table]{skip=0pt} 
	
	\thispagestyle{empty}

\begin{titlepage}
\begin{flushleft}

{\MakeUppercase{\bfseries An evaluation of information sharing parking guidance policies using a Bayesian approach}}\\[36pt]

{\bfseries Xinyi Wu} \\
Industrial and Operations Engineering \\
University of Michigan\\
1205 Beal Avenue \\
Ann Arbor, MI 48109-2117 \\
Tel: 734-846-1763 \hspace{0.1in}  Email: xinyiwu@umich.edu\\[12pt]

{\bfseries Kartik Balkumar}\\
Industrial and Operations Engineering \\
University of Michigan\\
1205 Beal Avenue \\
Ann Arbor, MI 48109-2117 \\
Tel: 201-625-5169 \hspace{0.1in} Email: kartikbk@umich.edu \\[12pt]

{\bfseries Qi Luo (corresponding author)}\\
Industrial and Operations Engineering \\
University of Michigan\\
1891 IOE Building, 1205 Beal Avenue \\
Ann Arbor, MI 48109-2117 \\
Tel: 734-395-5781   \hspace{0.1in} Email: luoqi@umich.edu\\[12pt]

{\bfseries Robert Hampshire}\\
University of Michigan Transportation Research Institute \\
2901 Baxter Road \\
Ann Arbor, MI 48109-2150 \\
Tel: 734-763-7746 \hspace{0.1in} Email: hamp@umich.edu\\[12pt]

{\bfseries Romesh Saigal}\\
Industrial and Operations Engineering \\
University of Michigan\\
2883 IOE, 1205 Beal Avenue \\
Ann Arbor, MI 48109-2117 \\
Tel: 734-763-7544 Email: rsaigal@umich.edu\\[12pt]

Word Count:  3761 words + \total{figure} figure(s) + \total{table} table(s) = 5511~words\\[12pt]

Submission Date: \today
\end{flushleft}
\end{titlepage}


\newpage
\section{Abstract}
Real-time parking occupancy information is critical for a parking management system to facilitate drivers to park more efficiently. Recent advances in connected and automated vehicle technologies enable sensor-equipped cars (probe cars) to detect and broadcast available parking spaces when driving through parking lots. In this paper, we evaluate the impact of market penetration of probe cars on the system performance, and investigate different parking guidance policies to improve the data acquisition process. 
We adopt a simulation-based approach to impose four policies on an off-street parking lot influencing the behavior of probe cars to park in assigned parking spaces. This in turn effects the scanning route and the parking space occupancy estimations. The last policy we propose is a near-optimal guidance strategy that maximizes the information gain of posteriors.   
The results suggest that an efficient information gathering policy can compensate for low penetration of connected and automated vehicles.
We also highlight the policy trade-off that occur while attempting to maximize information gain through explorations and improve assignment accuracy through exploitations. Our results can assist urban policy makers in designing and managing smart parking systems.

\newpage

\section{Introduction}
The ever-increasing number of cars on roads today has led to a burden on the management of transportation infrastructure. A prime example we observe today is vehicle owners finding it difficult to search for parking spaces. This is more evident in large cities and in prime locations where it is not uncommon to find vehicles moving around inside a parking lot anticipating a parking space to free up. In addition to the driver discomfort and frustration, searching for a parking space leads to a significant loss of personal time and an increase in fuel consumption.

Searching for a parking space in an optimized manner is thus a problem that demands the attention of researchers. 
The availability of real-time parking information has the potential for immense time and economic savings as drivers know in advance the presence of empty parking spots at the end of their trips. A highly desirable feature in such a system is that the source of real-time information must be independent of infrastructure. Parking solutions that depend on infrastructure, for example the use of on-site cameras, often incur significant costs. More importantly the implementation of such a solution is not generic with respect to the layout of the parking lot. Modern vehicle advancements have led to cars that are today equipped with vision and range-based navigation sensors and offer a source of real-time parking space information which is independent of surrounding infrastructure.  The focus of researchers has thus shifted towards modifying the navigating sensors and algorithms to tap real-time information on the occupancy of parking spaces, and subsequently relaying the useful information to a wide audience \textit{\cite{Luo2016}}.

Even though exploiting data from vehicle sensors for generating parking information is now close to a reality, research on parking strategies and policies that maximize the quality of this real-time information system is still nascent. Hence this paper aims to bridge the research gap by introducing a simulation-based approach to develop and investigate optimal parking policies for an intelligent parking system that depends on information collected from vehicle sensors. We propose a parking simulator that simulates a real parking lot in the city of Ann Arbor in Michigan. Cars entering and leaving the lot are modelled as probe cars (equipped with sensors) and non-probe cars (without sensors) to implement a ``connected vehicle ''   environment. The developed policies are evaluated using the simulator based on the accuracy of occupancy prediction and a single optimal policy is determined.

The paper is structured as follows: i) Section II reviews related work on existing parking search models and optimal path planning in parking. ii) Section III describes the developed parking simulator comprised of four modules, and provides a description on the optimal parking space allocation policy for probe cars to maximize the quality and availability of real-time parking information iii) Section IV offers a visualization of the parking simulator and a comparison of results between parking information collected one-way and two-ways by probe cars. iv) Section V provides a summary of the current work and a final conclusion.

\section{Related Work}
We discuss previous studies in literature on parking search optimization and related works. Luo et al. proposed a parking detection algorithm based on data collected from range-based sensors on vehicle, and improved its accuracy by introducing SLAM method \textit{\cite{Luo2016}}. Most of input data and configurations in simulation are based on this project. Bogoslavskyi et al. proposed a Markov Decision Process (MDP) based planner to calculate paths that minimize the time it takes to search for a parking space and walking up to a target destination after parking \textit{\cite{Bogoslavskyi2015}}. They calculated the paths using occupancy probabilities of parking spaces. The occupancy probabilities are considered uncertain and derived from visual sensor data and prior probability estimates of the spaces. Farkas and Lendak used simulation to study the improvement in parking search cruise time from crowd-sensing real-time parking information in an urban environment \textit{\cite{Farkas2015}}. Real-time information on occupancy of parking spaces alone may not improve parking search time for drivers. For example, Tasseron et al. used simulation to understand the impact of disseminating on-street parking information using vehicle-vehicle communication and vehicle-sensor communication. They concluded that, in contrary to theoretical expectations, the cruise time for searching for parking spots does not decrease significantly and may also increase, even under occupancy rates as high as 90\% to 95\%. The authors link it to the more likelihood of drivers parking their cars before reaching their destinations \textit{\cite{Tasseron2015}}. In another undesirable situation, broadcasting real-time information on parking spaces can negatively affect system performance by impacting driver behavior in unexpected ways. Wahle et al. discussed several studies on the various negative impacts of sharing real-time parking space information with too many drivers \textit{\cite{Wahle2002}}. Under these scenarios, our study assumes that parking policies can be used as a medium to control parking probabilities and the undesirable effects of excess information while achieving reduction in parking search times for the entire system of vehicles.

Agent-based simulation models that involve parking space detection must be based on suitable route choices for probe cars in order to maximize the sensing of the parking spaces and maintain latest information. Several models in the fields of robotics and computer science have dealt with this type of path planning problem. Singh et al. proposed an efficient algorithm for the Multi-robot Informative Path Planning Problem (MIPP) to generate informative paths while maximizing a sub-modular function like mutual information \textit{\cite{singh2007efficient}}. Martinez-Cantin et al. proposed a Bayesian optimization method for a mobile robot planning its path for optimal sensing of the environment under time constraints  \textit{\cite{martinex2009}}. Chekuri et al. set up an ``orienteering problem'' for a weighted and directed graph. Nodes of the graph are visited by a walk and the algorithm developed maximizes a submodular set function associated with the nodes visited \textit{\cite{chekuri2005}}.

\section{Parking Simulator based on Connected Vehicles Technology}
The simulator consists of three modules, event module, routing module and scanning module, whose output are connected to a data visualizer. The event module generates consequent arrivals and departures of both probe cars and normal cars, and each car will follow the route decided by the routing module. Finally, the scanning module will be activated only for probe cars and the estimated states of parking spaces along its trip will be updated using Bayes rule. All three modules transmit their status to the visualization module simultaneously.   

\subsection{Event Module}
Consequent events in the event module include arrivals and departures of probe cars (type 1 car) or normal cars (type 2 car). It is reasonable to formulate the parking arrival/departure process as a $M_t/M/N/C$ queue as a service system, where $N$ servers represents $N$ parking spaces respectively, and $C$ is the queue capacity, i.e. maximum number of cars waiting in the parking lot for next available parking space. Arrival process's intensity is $\lambda(t)$ as a function of time during the day, and the type of the next arriving car is randomly decided with fixed ratio $\gamma$. The arrival intensity follows a piecewise function in units of cars per hour. It is set to emulate the pattern of intra-city traffic with average rate of 120 vehicles per hour, whose intensity is much higher in early morning and late afternoon when people drive in and out for the regular work hours, and slightly higher at noon when people drive during their lunch break \textit{\cite{geroliminis2009}}. 

\begin{align*}
    & \lambda(t) = 
    \begin{cases}
        288 & \text{ if } 0 \leq t < 1 \text{ or } 8 \leq t < 9 \\
        72 & \text{ if } 1 \leq t < 3 \text{ or } 7 \leq t < 8 \\
        144 & \text{ if } 4 \leq t < 6 
    \end{cases}
\end{align*}

Each arriving car will be assigned to a parking space according to a specified policy in the routing module unless the number of cars in the system $n > N + C$. The service completion time (time spent in the parking lot) are exponential variable with constant parameter $\mu$. First come first serve rule is applied in the queue when all parking spaces are full, and cars arrive when the queue buffer has reached the capacity will leave the system immediately.

The System State ($SS$) variable is a vector of $(N+2)$ dimensions. The first $N$ elements are Boolean variable indicating the occupancy of $N$ parking spaces accordingly. The $N+1$ element $n$ represents the total number of cars in the system, and the last element $c$ indicates the number of cars in the queue. Counter variables $N_A$ are total number of arrivals of probe cars / normal cars by time $t$, and $N_D$ are total number of departures by time $t$. By tracking these two set of variables, we are able to observe how the system evolves with time by generating a discrete event list, which can be referred in the general parallel servers queue simulations in \textit{\cite{ross2013}}. 
 
The discrete event simulation that generates an event list is presented in the pseudocode below:

\algblock[Name]{Start}{End}
\begin{algorithm}[H]
\caption{Event Module}
\begin{algorithmic}
\State \textbf{Variables:}
\State  $TT$: Total time of simulation
\State  $SS=(X_1,\cdots,X_N,n,c)$
\State  Actual status of parking space $i$ $X_i=\begin{cases}
 0 & \text{if the parking space $i$ is free}\\
 1 & \text{if the parking space $i$ is occupied by a probe car}\\
 2 & \text{if the parking space $i$ is occupied by a normal car}
 \end{cases}
 $
\State \textbf{Initialization:}
\State $t=N_{A,1}=N_{A,2}=N_{D,1}=N_{D,2}=0$
\State $X_i = 0, i = 1, \cdots, N$
\State $n = c = 0$
\State Generate $T_0$ and indicating variable $z_0$. If $z_0 == 1$, set $t_{A,1}=T_0$, $t_{A,2}=0$; else, set $t_{A,1}=0$, $t_{A,2}=T_0$. 
\State $t_i = \infty, i = 1, \cdots, N$
\Start
    \While{$t<TT$}
        \If{$t_{A,1}$==min\{$t_{A,1},t_{A,2},t_i$\}} \Comment{Arrival of Probe Car}
            \State  $t = t_{A,1}$
            \State $N_{A,1} = N_{A,1}+1$
            \State Generate $T_t$ and reset $t_{A,1}=t_{A,1}+T_t$.
            \If{$n<N$} 
                \State  Assign a parking space $I$ according to the parking policy.
                \State  $n = n+1$
            \algstore{Event Module}
\end{algorithmic}
\end{algorithm}

\begin{algorithm}[H]
\begin{algorithmic}
    \algrestore{Event Module}
    \State (continued)
                \State  $SS=(0,\cdots,X_I = 1,\cdots,0,n,c)$
                \State  Generate $Y$ and reset $t_I = t + Y$.
                \State  Activate the routing and scanning models.
            \ElsIf{$n<N+C$}
                \State  $n = n+1$
                \State  $c = c+1$
                \State  $SS=(X_1,\cdots,X_N,n,c)$
                \State  Activate the routing and scanning models.
            \EndIf
        \ElsIf{$t_{A,2}$==min\{$t_{A,1},t_{A,2},t_i$\}} \Comment{Arrival of Normal Car}
            \State  $t = t_{A,2}$
            \State  $N_{A,2} = N_{A,2}+1$
            \State  Generate $T_t$ and reset $t_{A,2}=t_{A,2}+T_t$.            
            \If{$n<N$}
                \State  Assign a parking space $I$ according to the parking policy.
                \State  $n = n+1$
                \State  $SS=(0,\cdots,X_I = 2,\cdots,0,n,c)$
                \State  Generate $Y$ and reset $t_I = t + Y$  
            \ElsIf{n<N+C}.
                \State  $n = n+1$
                \State  $c = c+1$
                \State  $SS=(X_1,\cdots,X_N,n,c)$
            \EndIf
        \ElsIf{$t_I$==min\{$t_{A,1},t_{A,2},t_i$\}, $i = 1, \cdots, N$} \Comment{Departure}
            \State  $t = t_I$
            \If{$X_I==1$} \Comment{Departure of Probe Car}
                \State $N_{D,1} = N_{D,1} + 1$
                \State Activate the routing and scanning models.
            \ElsIf{$X_I==2$} \Comment{Departure of Normal Car}
                \State $N_{D,2} = N_{D,2} + 1$
            \EndIf
            \State $X_I = 0$
            \State $t_I = \infty$
            \If{$n>N$}
                \State $X_I = X_{N+1}$
                \State Generate $Y$ and reset $t_I = t_I + Y$.
                \For{$i$ for $N+1$ to $n-1$}
                    \State $X_i = X_{i+1}$
                \EndFor
                \State $X_n = 0$
            \EndIf
        \EndIf
    \EndWhile
\End
\end{algorithmic}
\end{algorithm}

\subsection{Routing Module}
One major objective of this simulation research is to find the impacts of different parking assignment policies on the service performance, which are integrated in the routing module. This module has two functions. One is to assign an available parking space to the arrived car, and the other is to generate driving routes for probe cars when they arrive or depart, which will be used in the scanning model later. For sake of simplifying workflow, the routing module should not intervene the event module as a precedent module, thus we assume that an arriving vehicle instantaneously travel to their parking locations. 

We evaluate four policies for assigning cars to parking spaces: 
\begin{itemize}
    \setlength\itemsep{0em}
    \item[- Policy 1] Random assignment: both types of cars are assigned to available parking spaces randomly, which is expected to represent the average performance of the system.
    \item[- Policy 2] Nearest parking: both types of cars park to the available parking space closest to the entrance. This is to simulate a destination-oriented parking policy assuming the entrance is the final destination for all drivers.
    \item[- Policy 3] Maximum satisfaction guidance: normal cars park to the space closest to the entrance while probe cars park to the space which is estimated to be most likely empty. In other words, this is to function as the maximum exploitation policy. 
    \item[- Policy 4] Near-optimal guidance: normal cars park to the space closest to the entrance while probe cars park to the space which will maximize information gain from scanning. In other words, this is to function as the maximum exploration policy. 
\end{itemize}

For generating driving routes, cars always arrive in the shortest path but there are two policies for departure routes. One is that the roads in the parking lot are two-way so cars follow the same path as they arrive. The other is that the roads are one-way so cars follow a different path when leaving. For an unregulated parking lot, the direction of cars driving in and out is always a mix of both routes so it is sufficient to research only these two extreme cases.

\subsubsection{A Near-Optimal Guide Policy (Policy 4)}
For each arriving car at time $t$, the action $a(t)$ consists of assigning to a parking space $i$  and choosing the route on a vertex-edge graph. In pursuance of improving the quality of parking guidance services, the principal goal is to accelerate the exploration process by distributing probe cars optimally. A very natural way to quantify the informativeness of a chosen route is mutual information, more specifically, the information gain from scanning vertex (parking spaces) of route. We only consider the posterior of each action because of the instantaneous routing assumption, thus the entropy is measured over the vertex along the route. 

\begin{equation*}
MI(\chi_N(t), a(t)) = H(\chi_N(t)) - H(\chi_N(t) | \chi_{a(t)} ).
\end{equation*}

where $\chi_i, i = 1, \cdots, N$ represents the estimated status of parking space $i$, whose measure is between 0 and 1. $H(\chi_N(t))$ is the entropy of the all vertex, and $ H(\chi_N(t) | \chi_{a(t)})$ is the conditional entropy by taking action $a(t)$ representing the observed set. Thus the mutual information measures the uncertainty reduction resulting from a chosen route. Chekuri et al. shows that $MI()$ is a submodular function, which possesses diminishing return property: the more locations already been sensed, the less information gained by sensing new locations \textit{\cite{chekuri2005}}. Furthermore, as the principle of optimality is strictly obeyed in sequence of posterior of scanning, we can find optimal policy for each probe car by maximizing $MI(\chi_N(t), a(t))$ by using direct policy search on the grounds of these facts:
\begin{enumerate}
    \setlength\itemsep{0em}
    \item For cases that $\gamma < 1$, i.e. a mixed arrivals of probe cars and normal cars, seeking for optimal policy becomes a NP-hard Partial Observable Markov decision process (POMDP) because the actual state $X_i$ is only known for those occupied by probe cars.
    \item The controllable objects are only probe cars, whose departure time is also a random variable since the profile of its departure path is stochastic.
    \item The action space is a high-dimensional compound matrix of routing module and scanning module.
\end{enumerate}

Inasmuch as it is prohibitive to find an explicit optimal policy, we implement an alternative approximation method using direct policy search at each step especially in the case that the candidate policy space is relatively small. In practice of robotics path planning \textit{\cite{singh2007efficient}}, this is proved to an efficient near-optimal policy to maximize information gain.

\subsection{Scanning Module}
The scanning module updates the estimated system state. When a probe arrives or departs, the scanning module is activated. We use recursive Bayesian updating in this module.  

\subsubsection{Recursive Bayesian Updating}
Let $\chi_{i,t}$ be a random variable of parking space $i$ at time $t$. The estimated state of parking space $i$ at time $t$ is $\hat{X}_i$, which is determined by

\begin{align*}
\hat{X}_i &= \begin{cases}
0  & \text{ if }  P(\chi_i) < 0.4 \\
1  & \text{ if }  P(\chi_i) > 0.6 \\
\text{null} & \text{ if }  0.4 \leq P(\chi_i) \leq 0.6  
\end{cases}.
\end{align*}

\noindent We are only interested in cases when $\hat{X}_i$ is not null. 

\begin{align*}
    \hat{X}_{i,t}&= \begin{cases}
    0 & \text{parking space $i$ is estimated to be empty at time $t$}\\
    1 & \text{parking space $i$ is estimated to be occupied at time $t$}
    \end{cases}.
\end{align*}
According to Bayes Theorem, 

\begin{align*}
    &P(X_i = 0|\hat{X}_i)= \frac{P(\hat{X}_i|X_i = 0)P(X_i = 0)}{P(\hat{X}_i)}\\
    &P(X_i \ne 0|\hat{X}_i)= \frac{P(\hat{X}_i|X_i \ne 0)P(X_i \ne 0)}{P(\hat{X}_i)}\\
    &P(\hat{X}_i) = P(\hat{X}_i|X_i = 0)P(X_i = 0)+P(\hat{X}_i|X_i \ne 0)P(X_i \ne 0).
\end{align*}

After each measurement, we use Bayes Theorem to compute the posterior probability $p_{i,t+1}$ that the parking space $i$ is occupied. The likelihood data are from the field tests in previous research, we conducted field tests to test the effectiveness of radar for detecting occupancy of parking spaces \textit{\cite{Luo2016}}. From this experience, we calculated the prior and likelihood data in Table \ref{table_bayesian_1} and Table \ref{table_bayesian_2}.

If the measurement is $\hat{X}_{i,t}=0$, then
\vspace{-0.05in}
\begin{table}[H]
\captionsetup{justification=raggedright,singlelinecheck=false}
\caption{Bayesian Updating for $\hat{X}_{i,t} = 0$}
\label{table_bayesian_1}
\begin{center}
\begin{tabular}{|c|c|c|c|}
    \hline
    Hypothesis & Prior $P(X_i)$ & Likelihood $P(\hat{X}_{i,t}=0|X_i)$ & Unnormalized Posterior \\
    \hline
    \hline
    $X_i \ne 0$ & $p_{i,t}$ & $0.093$ & $0.093 p_{i,t}$ \\
    \hline
    $X_i = 0$ & $1-p_{i,t}$ & $0.941$ & $0.941 (1-p_{i,t})$ \\
    \hline
\end{tabular}
\end{center}
\end{table}
\vspace{-0.25in}
\noindent If the measurement is $\hat{X}_{i,t}=1$, then
\vspace{-0.05in}
\begin{table}[H]
\captionsetup{justification=raggedright,singlelinecheck=false}
\caption{Bayesian Updating for $\hat{X}_{i,t}=1$}
\label{table_bayesian_2}
\begin{center}
\begin{tabular}{|c|c|c|c|}
    \hline
    Hypothesis & Prior $P(X_i)$ & Likelihood $P(\hat{X}_{i,t}=0|X_i)$ & Unnormalized Posterior \\
    \hline
    \hline
    $X_i \ne 0$ & $p_{i,t}$ & $0.907$ & $0.907 p_{i,t}$ \\
    \hline
    $X_i = 0$ & $1-p_{i,t}$ & $0.059 $ & $0.059 (1-p_{i,t})$ \\
    \hline
\end{tabular}
\end{center}
\end{table}
\vspace{-0.25in}
\noindent Therefore the posterior at time $t+1$ can be expressed as:

\begin{align*}
    \begin{cases}
         p_{i,t+1} = \frac{0.093 p_{i,t}}{0.093 p_{i,t}+0.941 (1-p_{i,t})} = \frac{0.093 p_{i,t}}{0.941 -0.848 p_{i,t}} & \text{ if } \hat{X}_{i,t}=0 \qquad (*) \\
         p_{i,t+1} = \frac{0.907 p_{i,t}}{0.907 p_{i,t}+0.059 (1-p_{i,t})} = \frac{0.907 p_{i,t}}{0.0589 +0.848 p_{i,t}} & \text{ if } \hat{X}_{i,t}= 1 \qquad (**)
    \end{cases}.
\end{align*}

\noindent The scanning module using recursive Bayesian updating is presented in the pseudocode below:

\begin{algorithm}[H]
\caption{Scanning Module}
\begin{algorithmic}
\State \textbf{Variables:}
\State $ES=(p_1, \cdots, p_N)$, in which $p_i$ represents the probability that parking space $i$ is occupied, $i = 1, \cdots, N$
\State \textbf{Initialization:}
\State $ES = (0.5, \cdots, 0.5)$
\Start
    \For{all parking spaces $i$ scanned by a probe car}
        \If{$\hat{X}_i == 0$}
            (*)
        \ElsIf{$\hat{X}_i == 1$}
            (**)
        \EndIf
    \EndFor
    \If{a probe car arrives and parks to parking space $I$}
        \State $p_I = 1$
    \EndIf
    \If{a probe car leaves parking space $I$}
        \State $p_I = 0$
    \EndIf
\End
\end{algorithmic}
\end{algorithm}

The updating of estimated matrix $\chi$ also includes a discount factor $\beta$ since the perceived parking states will diminish to unknown status ($p = 0.5$) at each time step. For a parking space $i$ that was scanned at $t$ but not scanned during $\Delta t$, 

\begin{align*}
    \chi_i (t+ \Delta t) = 
    \begin{cases}
        0.5 + \beta^{\Delta t} (\chi_i(t) - 0.5)  & \text{ if } \chi_i(t) > 0.5\\
        0.5 - \beta^{\Delta t} ( 0.5 - \chi_i(t)) & \text{ if } \chi_i(t) < 0.5
    \end{cases}.
\end{align*}

\noindent Calculating the error of estimation at parking space $i$ following this operator

\begin{align*}
    e_i =
    \begin{cases}
        \hat{X}_i \oplus X_i & \text{ if } \hat{X}_i = 0 \text{ or } 1 \\
        1 & \text{ if } \hat{X}_i = \text{ null }
    \end{cases}.
\end{align*}
The total absolute error of all parking spaces is therefore

\begin{equation*}
    e = \sum_{i\in N} e_i.
\end{equation*}

\section{Results and Discussion}
\subsection{Visualization}
\begin{figure}[H]
    \centering
    \bf\subfloat[Field test parking lot with grid labels \textit{\cite{Luo2016}}.\label{subfig-1}]{%
      \includegraphics[width=0.66\textwidth]{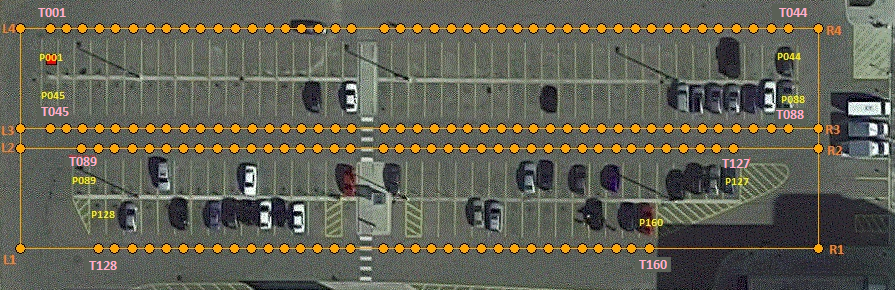}
    }
    \hfill
    \bf\subfloat[Parking simulator visualization.\label{subfig-2}]{%
      \includegraphics[width=0.7\textwidth]{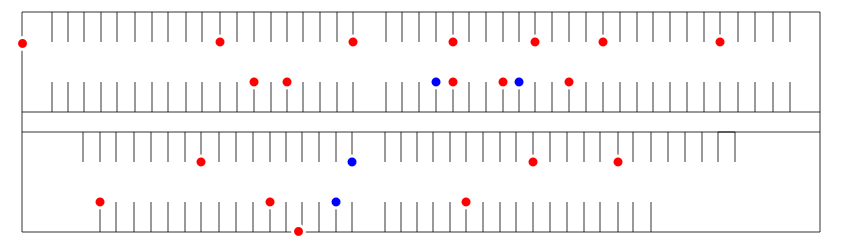}
    }
    \caption{Visualization module.}
    \label{figure snapshot}
\end{figure}

The visualizer is developed as a back-end python and a front-end web module. The visualizer is interfaced with the other simulation modules and illustrates the guided movement of probe cars and non-probe cars through the parking lot. The movement of cars is overlaid on an aerial image of the parking lot used in field test in \textit{\cite{Luo2016}}. The parking lot, with 160 parking spaces, is converted into a node-and-edge graph as shown in Figure \ref{figure snapshot}. Each node represents a parking space and the edges connecting the nodes are paths for cars to navigate inside the parking lot. Differing color schemes are used for the vehicles-red for probe cars and blue for non-probe cars. 

\subsection{Simulation Results}
\subsubsection{Simulation Parameters}
  Parameters of the simulator's modules are set as followed. In the event module,  arrival events are assumed to be a non-homogeneous Poisson process with arrival rate as a piecewise function of time in minutes. The parking time follows an exponential distribution with $\mu = 60$. In the scanning module, the initial state of all parking spaces are empty with estimation $p(\chi_i) = 0.5, i = 1,\cdots, N$. The discount factor of estimation is $\beta=0.9$. The scanning range of a probe car is at most 6 nodes surrounding the current position.
 
 We use the relative error of posteriors to quantify the performance of each policy. The relative error between the predicted occupancy and actual occupancy is calculated for increasing market penetration of probe cars in percentage, which is defined as: 

\begin{align*}
    e(t) = \frac{\text{number of wrongly estimated parking space}}{\text{number of total parking spaces}}.
\end{align*}
The average error of a given event list is calculated over $t^*$, which are consequent events in the list:

\begin{align*}
    \bar{e} = \frac{\int_0^T e(t) dt}{T} = \frac{\sum_{t^*} e(t^*) \Delta t}{T}.
\end{align*}

\subsubsection{Relative error changing with time}
\begin{figure}[H]
\begin{center}
    \includegraphics[scale=0.38]{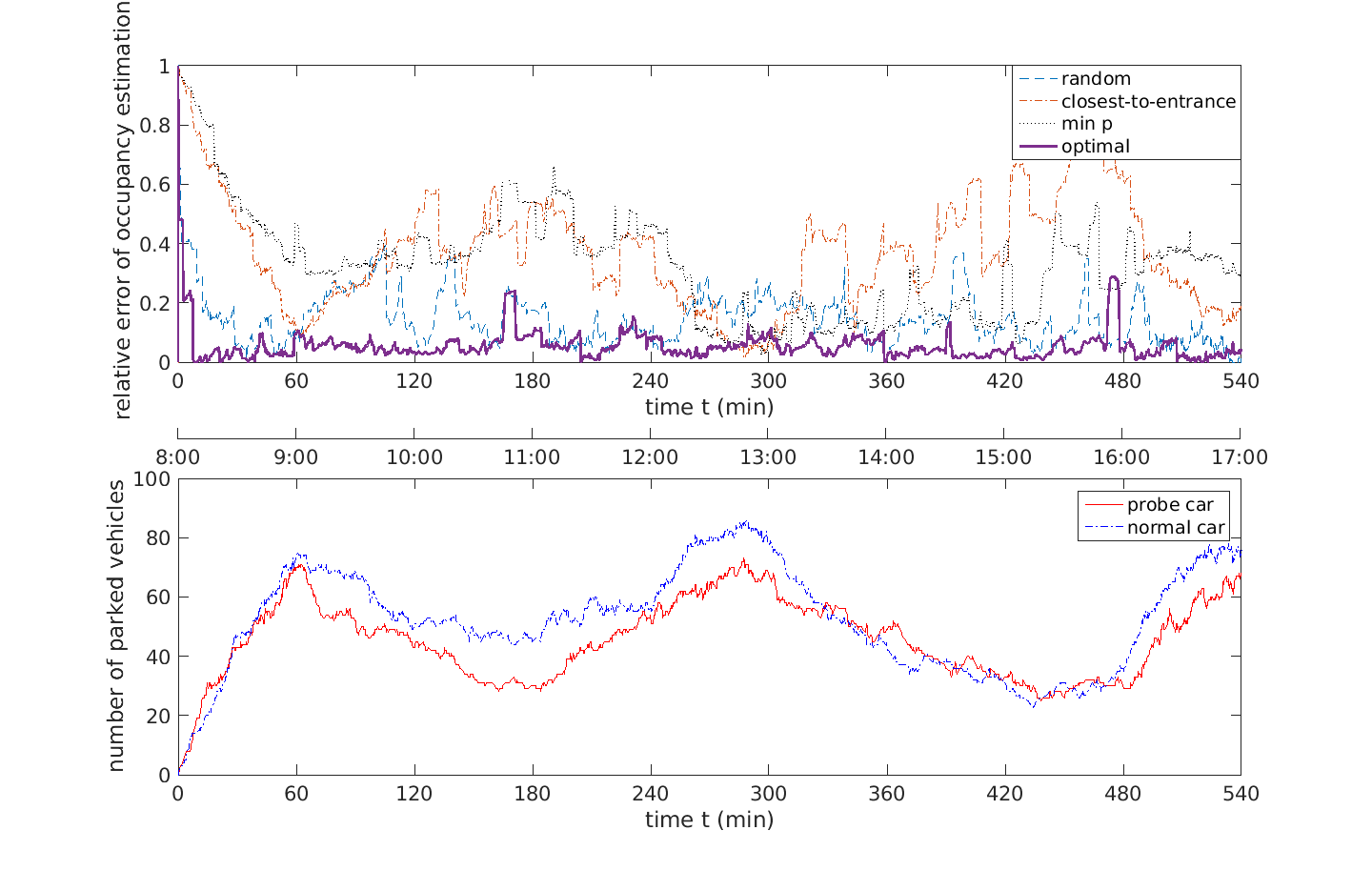}
    \caption{Two-way parking simulation result (percentage of probe car $\gamma = 0.5$).}
    \label{figure_errorwithtime_twoway}
\end{center}
\end{figure}

\begin{figure}[H]
\centering
    \includegraphics[scale=0.38]{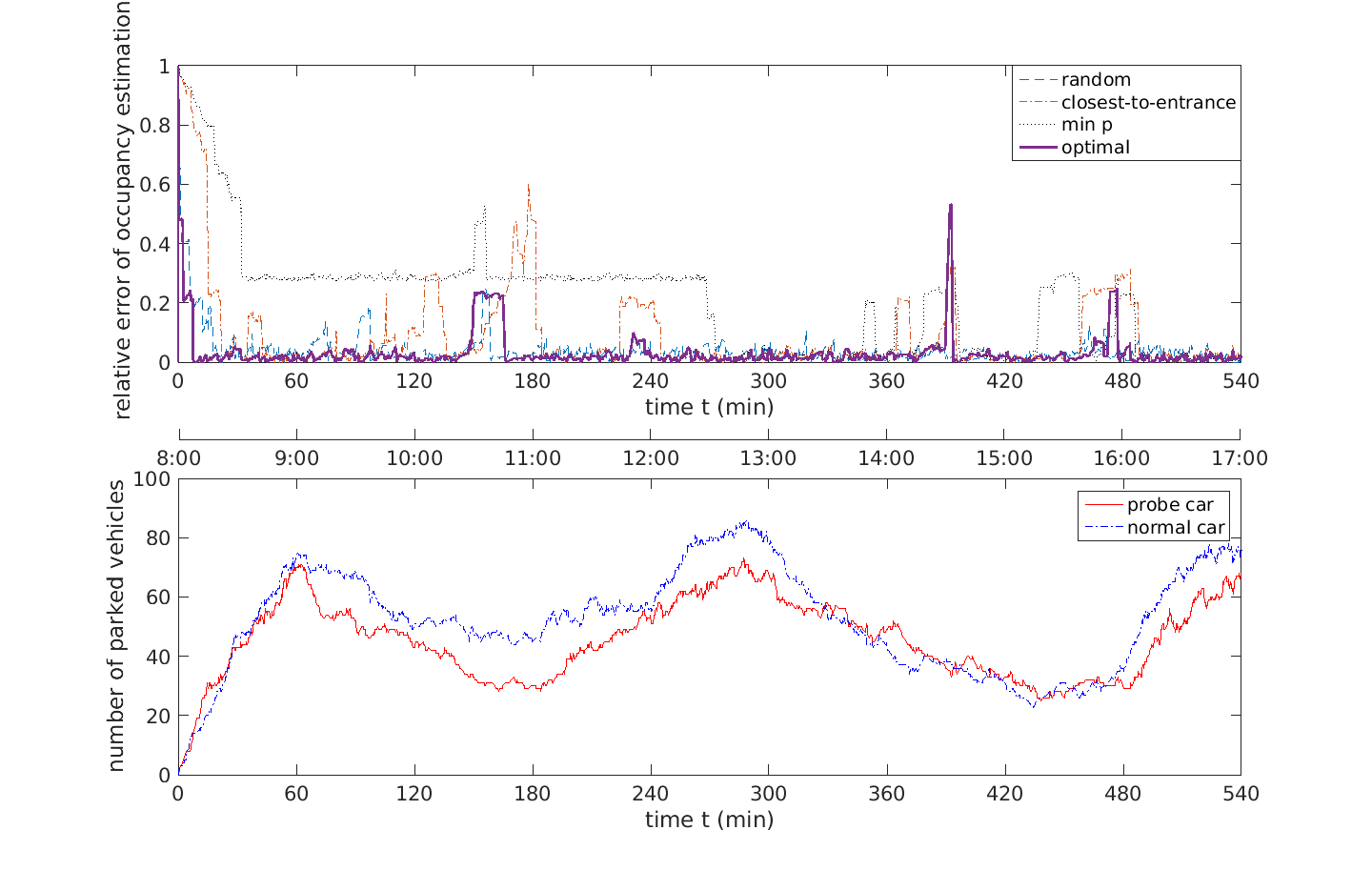}
    \caption{One-way parking simulation result (percentage of probe car $\gamma = 0.5$).}
    \label{figure_errorwithtime_oneway}
\end{figure}

Since the arrival rate and number of vehicles in the parking lot change with time, the relative error fluctuate as a function of system state. Their relationship is shown in Figure \ref{figure_errorwithtime_twoway} of one-way parking and Figure \ref{figure_errorwithtime_oneway} of two-way parking both with the percentage of probe cars in arrivals $\gamma=0.5$. 

The flow of arriving probe cars and normal cars in the simulation follows the pattern of typical daily intra-city traffic in literature \textit{\cite{geroliminis2009}}. In both cases, at zero time the relative error is large across all the policies because of the initial conditions, but tends to drop considerably as information is gained, and keeps oscillating as information variation occurs. Notice that the near-optimal guide policy is stable over time in contrast to the other policies whose estimation errors fluctuate over time. In addition this developed policy is less sensitive to the number of parked vehicles in the system while the errors of others increase when the number of parked vehicles drops. The very few peaks in its errors are caused by the diminishing of information, seeing that there are no probe cars arriving during that time and estimators $\chi_i$ fall back into the undecided range.    

Another observation is that the oscillation of errors in one-way case is relatively smaller in size than in two-way case for all policies. This is because one-way path allows probe cars to scan towards different direction from the route their arrived. Ideally, if there is no discount factor in estimations, choosing any parking space in the same row has equivalent effects on the error means. 

\subsubsection{Relative error of different parking policies}

\begin{figure}[H]
    \centering
    \includegraphics[scale=0.55]{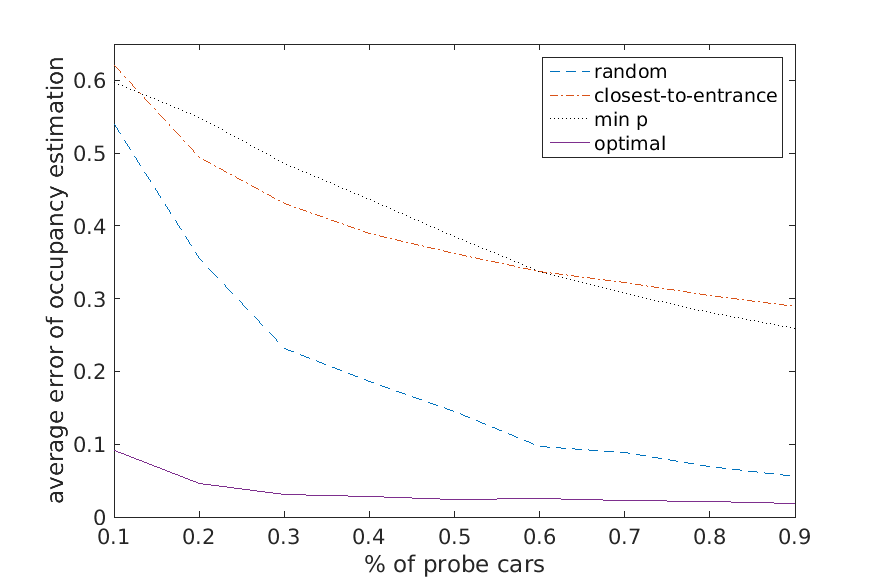}
    \caption{Two-way parking simulation result.}
    \label{figure_twoway}
\end{figure}

\begin{figure}[H]
    \centering
    \includegraphics[scale=0.55]{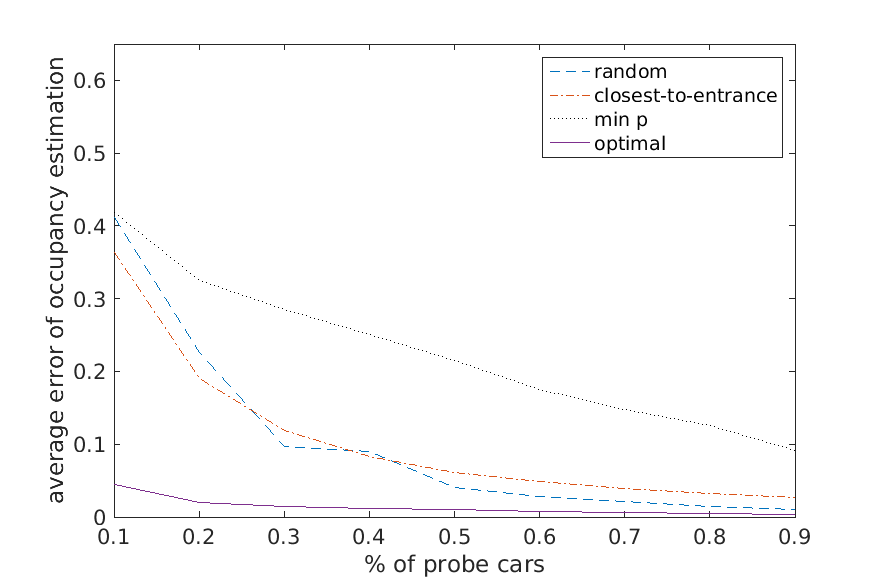}
    \caption{One-way parking simulation result.}
    \label{figure_oneway}
\end{figure}

In order to observe the average performance of each policy, the Monte Carlo method is applied to generate event list repeatedly. The percentage of probe cars in arrivals $\gamma$ ranges from 0.1 to 0.9 with step 0.1. The simulation is repeated for 1,000 times for each $\gamma$. The average relative error of estimations with increasing percentage of probe cars in the arrivals are of two-way parking is shown in Figure \ref{figure_twoway}, and the result of one-way parking is shown in Figure \ref{figure_oneway}.

The near-optimal parking policy outperforms in both cases in average. On the contrary, the high satisfaction policy which assigning cars to most probably available parking space is always modest in contributing to the estimations. This result is intuitive because over-exploiting the given information will sacrifice exploring the unknown regions. This trade-off has to be considered through the decision-making which is able to improve the parking guidance and also guarantee to satisfy individual drivers' demands.   

Another important finding is that using the near-optimal policy can significantly compensate for low fleet penetration of probe cars. As the result shows, the performance of near-optimal policy is acceptable even with low penetration of probe cars, and the relative error can only be slightly reduced when the percentage increases from 10\% to 90\%. According to the market prediction in literature \cite{Luo_market2016}, even by assuming that the growth of ADAS market follows the logistic function, the penetration rate of probe cars will be less than 20\% by 2020, which is not adequate for other policies.
Therefore we can conclude that applying the near-optimal policy will enable covering the parking lot with low percentage of cars equipped with sensors.  On the other hand we should not overstate the effect because the assumption that the near-optimal policy always directs drivers to available parking spaces. In practice, there is a chance that our policy directs drivers to a space that is actually not available. In future work this adverse event can be incorporated into the analysis.

\section{Conclusion}
This paper described how we build a parking simulator consisting of event module, routing module, scanning module and visualizer. Four parking guidance policies are tested on this platform with mixed traffic of probe cars and normal cars. Comparing to the other policies, the near-optimal policy stably and accurately estimates the parking occupancy in the repeated experiments. A further discussion on the trade-off of exploitation and exploration in optimal routing provides insight for an improved parking guide policy for probe cars.    

Future work will focus on building a multistage stochastic programming model for parking. First part of the work is to formulate a cost function for optimal parking guidance policy including the misguide penalty. Second task is to extend the current one-stage policy to a multistage policy in scenarios that the arriving car is assigned to a occupied parking space. This will require the compound modeling of generating events and routing. These improvements of simulation will be able to give insight into a more comprehensive exploration-exploitation parking guidance policy.      

\section{Acknowledgement}
This project was funded by Mobility Transformation Center.

\bibliographystyle{trb}

\newpage

\section{Appendix}
\nomenclature[01]{$N$}{Number of servers (parking spaces)}
\nomenclature[02]{$n$}{Number of cars in the system}
\nomenclature[03]{$C$}{Maximum number of cars waiting for parking}
\nomenclature[04]{$c$}{Number of cars in the queue}
\nomenclature[06]{$t$}{Time variable of the simulation}
\nomenclature[07]{$N_{A,1}$}{Number of probe cars arrivals by $t$}
\nomenclature[07]{$N_{A,2}$}{Number of normal cars arrivals by $t$}
\nomenclature[08]{$N_{D,1}$}{Number of probe cars departures by $t$}
\nomenclature[08]{$N_{D,2}$}{Number of normal cars departures by $t$}
\nomenclature[09]{$\lambda (t)$}{Arrival rate of cars, a function of time during the day}
\nomenclature[10]{$\mu$}{Average service completion rate}
\nomenclature[11]{$SS$}{System state variables}
\nomenclature[11]{$ES$}{Estimated system state variables}
\nomenclature[12]{$t_{A,1}$}{Time of the next arrival of probe car}
\nomenclature[12]{$t_{A,2}$}{Time of the next arrival of normal car}
\nomenclature[13]{$t_i$}{Service completion time for the car in sever $i$}
\nomenclature[14]{$A(n)$}{Arrival time of car $n$}
\nomenclature[15]{$D(n)$}{Departure time of car $n$}
\nomenclature[16]{$X_i$}{Actual status of sever (parking space) $i$}
\nomenclature[17]{$\chi_i$}{Measure of server (parking space) $i$}
\nomenclature[18]{$\hat{X}_i$}{Estimated status of server (parking space) $i$}
\nomenclature[18]{$\gamma$}{Ratio of probe cars in total traffic}
\nomenclature[19]{$\beta$}{Discount factor of estimation}
\nomenclature[20]{$a(t)$}{Action at time t, a combined decision of assignment and route of next arriving car}
\nomenclature[21]{$e$}{Error of estimation}
\printnomenclature

\end{document}